\documentclass[conference]{IEEEtran}
\usepackage[linesnumbered,ruled]{algorithm2e}
\usepackage{enumerate}
\usepackage{graphicx}
\newcommand\nop[1]{}
\def\mc{\ensuremath{\mathcal}}
\def\mi{\ensuremath{\mathit}}
\def\mr{\ensuremath{\mathrm}}
\newtheorem{theorem}{Theorem}
\newtheorem{example}[theorem]{Example}

\ifCLASSINFOpdf
\else
\fi
    \IEEEoverridecommandlockouts
    \IEEEpubid{\makebox[\columnwidth]{978-1-4673-5494-3/13/\$31.00~\copyright~2013 IEEE \hfill}
    \hspace{\columnsep}\makebox[\columnwidth]{ }}

\begin{document}
%
\title{Matching Demand with Supply in the Smart Grid using
Agent-Based Multiunit Auction}

\author{\IEEEauthorblockN{Tri Kurniawan Wijaya}
\IEEEauthorblockA{
\'{E}cole Polytechnique F\'{e}d\'{e}rale de Lausanne\\
Switzerland\\
Email: tri-kurniawan.wijaya@epfl.ch
}
\and
\IEEEauthorblockN{Kate Larson}
\IEEEauthorblockA{University of Waterloo\\
Canada\\
Email: klarson@uwaterloo.ca}
\and
\IEEEauthorblockN{Karl Aberer}
\IEEEauthorblockA{
\'{E}cole Polytechnique F\'{e}d\'{e}rale de Lausanne\\
Switzerland\\
Email: karl.aberer@epfl.ch
}
}


%


\maketitle

\begin{abstract}
%
Recent work has suggested reducing
electricity generation cost by cutting 
the peak to average ratio (PAR) without reducing the total amount of the loads.
However, most of these proposals rely on consumer's willingness to act.
In this paper, we propose an approach to 
cut PAR explicitly from the supply side.
The resulting cut loads are then distributed among 
consumers by the means of a multiunit auction 
which is done by an intelligent agent 
on behalf of the consumer.
This approach is also in line with 
the future vision of the smart grid to have 
the demand side matched with the supply side.
Experiments suggest that our approach reduces overall system cost and gives benefit to both consumers and the energy provider.
\end{abstract}



%
\IEEEpeerreviewmaketitle

\section{Introduction}



Currently, electricity markets are designed so that the electricity supply has to fulfill the demand. When the demand increases rapidly, several problems occur, such as the possibility of power failures and high generation costs as expensive generators are turned on to fulfill demand for short peak periods. One vision for the smart grid has been to reverse this, and instead have demand match the available supply~\cite{Ramchurn2012}.  
In this paper we propose a method towards achieving this end.


The key to our method is the explicit cut of the peak to average ratio (PAR) of the electricity load generated. We introduce an approach which cuts PAR and then lets consumers adapt by using an auction for redistributing the load. As others have proposed previously, automated intelligent agents can be used to represent consumers in these auctions, making the entire process seamless from the consumers' perspective~\cite{Oh2008,Ramchurn2011,Vytelingum2011}.


Both the challenges of obtaining low PAR and the advantages have been widely investigated~\cite{USDE06,Cappers10,Strbac2008}. 
However, to the best of our knowledge, no other work proposes 
to explicitly cut PAR and let the demand side adapt to the supply. 
Many other approaches which have investigated cutting PAR have 
focussed on having the demand side voluntarily adjust their 
consumption (given various incentives) in order to reduce 
peak load~\cite{Li2011,Mohsenian2010,Ganu2012}.



Using auctions in electricity markets has been both proposed in the literature and been used in practice, see for example~\cite{Bower2001},~\cite{Contreras2001}. In our work, however, we take a close look at the relationship between produces and consumers and realize the vision of matching demand with supply while also considering demand satisfiability and minimum load guarantees, while also supporting the possibility of additional benefits for each side.  
The work most similar to that proposed here looked at the use of service curves~\cite{Leboudec11}. This work is similar in that it also imposed certain restrictions on the consumers' consumption. However, in their work, the producers and consumers had to agree in advance on the service curves contract, reducing the ability to adapt to changing supply conditions.

This paper makes the following contributions:
\begin{itemize}
\item We propose an algorithm to explicitly cut PAR and prove its soundness and completeness.

\item We introduce a (multiunit) auction for distributing load after the PAR cut. We provide a minimum load guarantee for all consumers, and show that the auction support truthful myopic bidding.

\item Our simulations illustrate that our method benefits both consumers and producers.
\end{itemize}

The rest of the paper is organized as follows. We describe our model and other basic notions in Section II. In Section III we describe the algorithm for cutting PAR and prove that it is both sound and complete, while in Section IV we present our auction for distributing available load. Our experimental results are presented in Section V, which conclusions and further discussions in Section VI.

%


\section{Preliminaries}
\label{sec:prelim}
\subsection{Load Modeling}

\noindent
\textbf{Load.} Let $\mc{N}$ be a set of consumers 
and $\mc{T} = \{t_1, \ldots, t_{|\mc{T}|}\}$ be a set of 
uniform time slots in a day.
We denote the \emph{electricity load} (or simply \emph{load}) needed by consumer $i \in \mc{N}$
in time slot $t \in \mc{T}$ as $L_i(t)$.
Hence, the total load of consumer $i$ for a day is given by: 
\[ L_i = \sum_{t \in \mc{T}} L_i(t).\] 
In addition, we define total load per time slot over all consumers as: 
\[ L(t) = \sum_{i \in \mc{N}} L_i(t)\]

\noindent
\textbf{PAR.} Peak to average ratio ($\mi{PAR}$) is a commonly 
used measurement to express how the peak compares to the average load:
\begin{equation}
\label{eq:par}
\mi{PAR}(L) = \frac{|\mc{T}|}{\sum_{t \in \mc{T}} L(t)} \; \max_{t \in \mc{T}} L(t) 
\end{equation}

\subsection{Multiunit Auction}
\label{sec:auction}

Before introducing our auction (in Section~\ref{sec:multi}), in this section we briefly explain the basics of multiunit auctions.

\vspace*{0.5ex}
\noindent
\textbf{Auction.} We use a \emph{uniform price auction}, where 
each winning bidder pays the same price for each item/resource they win. 
The price paid is the price of the highest non-winning bidder.
%
Given $R$ resources and $\mc{N}$ as a set of agents or bidders,
we denote a \emph{bid} of an agent $i \in \mc{N}$ as a 2-tuple $(r_i, v_i)$ where $r_i$ is the number of resources desired by agent $i$ and $v_i$ is her \emph{valuation} (the price $i$ willing to pay for each resource she wins).

\vspace*{0.5ex}
\noindent
\textbf{Winners.} 
The resources are won by the $k$ highest bidders. 
Let $W \subseteq \mc{N}$ be the set of $k$ agents who win the auction, then we require that:

\begin{enumerate}[(i)]

\item the winners are the highest bidders: $\forall i,j \hspace{1ex} (v_i \geq v_j)$ where $i \in W$ and $j \in \mc{N} \setminus W$,

\item we allocate, tentatively, the maximum number of resources: 
if $W \subset \mc{N}$ then $\big(\sum_{i \in W} r_i \big) \geq R$, and

\item $W$ is the smallest set of winners: 
$\forall W' \; \big( \sum_{i \in W'} r_i \big) < R$ where $W' \subset \mc{N}$.

\end{enumerate}

\vspace*{0ex}
\noindent
\textbf{Price. } 
The price paid by each winner $i \in W$ is the valuation of the highest non-winning bidder, 
i.e., $p^* = v_{j}$ where $\forall j' (v_j \geq v_j')$ where  $j,j' \in \mc{N} \setminus W$. 
%
%
Note that when $W  = \mc{N}$ then there are two possibilities, either all bidders pay 0 or pay the reserve price (a price which is fixed by the auctioneer as the minimum price for a resource).

\vspace*{0.5ex}
\noindent
\textbf{Resource distribution.}
We sort the winning bidders by their valuation in ascending order.
The resources are distributed to the winners starting from the highest bidders. Hence, there could be the case where the lowest winning bidder gets resources less than what she desires. 
In this case, she has an option to walk away (cancel her participation in the auction) or accept the resources offered. 

\vspace*{1ex}
\begin{example}[Multiunit Auction]
A company would like to sell 6 resources. Then, 

\begin{itemize}
\item bidder 1 would like to buy 2 resources at \$12, 
\item bidder 2 would like to buy 3 resources at \$10,
\item bidder 3 would like to buy 3 resources at \$8, and
\item bidder 4 would like to buy 1 resources at \$6.
\item bidder 5 would like to buy 2 resources at \$5.
\end{itemize}

Hence the winners of the auction are bidder 1, 2, and 3. 
Bidder 1, 2, and 3 got 2, 3, and 1 item respectively 
where each of them have to pay \$6 for an item. 
In this case, since the total demand of bidder 3 is not met, 
she can decide whether to take the 1 item offered and pay \$6, 
or withdraw from the auction (pay nothing and receive no item).
\end{example}
\vspace*{1ex}


\section{PAR-Cut}
\label{sec:parcut}

We cut $\mi{PAR}(L)$ by a cut percentage, $c$, resulting in a new load vector $L'$:

\begin{equation}
\label{eq:cutpar}
(1-c) \cdot \mi{PAR}(L) = \mi{PAR}(L'),
\end{equation}

and

\begin{equation}
\sum_{t \in \mc{T}} L(t) = \sum_{t \in \mc{T}} L'(t).
\label{eq:fixLoadSum}
\end{equation}

\noindent
When condition in Eq.~\ref{eq:fixLoadSum} is met, 
using Eq.~\ref{eq:par} we can rewrite Eq.~\ref{eq:cutpar} as:
\begin{equation}
\label{eq:cutPeak}
(1-c) \cdot \max_{t \in \mc{T}} L(t) = \max_{t \in \mc{T}} L'(t).
\end{equation}

In Algorithm~\ref{alg:parcut} we provide a technique to explicitly cut the $\mi{PAR}$ of the original load generated by consumer demand. The returned result is a load whose $\mi{PAR}$ has been cut. 
Cutting $\mi{PAR}$ while maintaining overall amount of load (as described in Section~\ref{sec:parcut}) means that there are some amount of load shifted from their original time slot. 
In this algorithm we aim to minimize the shift distance by first 
attempting to shift to a neighboring time slot.

There are several helper methods used in Algorithm~\ref{alg:parcut}:
\begin{itemize}
\item $\mr{findPeak}(L)$: returns the peak load of a load vector $L$.
\item $\mr{min}(A,B)$: returns the smallest between two number $A$ and $B$.
\item $\mr{shift}(\mi{x}, L, t_1, t_2)$: 
moves $x$ amount of load from $L_{t_1}$ to $L_{t_2}$.
\end{itemize}

\vspace*{-1ex}
\begin{algorithm}
\label{alg:parcut}
\caption{PAR-Cut}
\DontPrintSemicolon
\SetAlgoVlined
\SetKwComment{Comment}{/*}{*/}
\KwIn{cut percentage $0 < c \leq 1$, load vector $L$}
$\mi{p'} \gets (1-c) \cdot \mr{findPeak}(L)$ \label{ln:newPeak}\;
let $\mc{T} = \{t_1, \ldots, t_{|\mc{T}|}\}$ \;
\ForEach{$t_i \in \mc{T}$}{
	\If{$L[t_i] > \mi{p'}$ \label{ln:check}}{
		$d \gets 1$ \label{ln:shiftStart}\; 
		$\mi{x} \gets L[t_i] - \mi{p'}$ \label{ln:excessLoad}\Comment*{excess load} 
		\While{$\mi{x} > 0$}{
			\If{$(i+d > |\mc{T}|) \wedge (i-d < 1) $}{ \label{ln:if1}
				\Return $\mi{fail}$ \label{ln:fail}
			}
			\If{$(i+d \leq |\mc{T}|) \wedge (L[t_{i+d}] < p')$ \label{ln:dAfter}}{
				$\mi{loadShifted} \gets \min(\mi{x},\mi{p'}-L[t_{i+d}])$\;
				$\mr{shift}(\mi{loadShifted}, L, t_i, t_{i+d})$\;
				$\mi{x} \gets \mi{x} - \mi{loadShifted}$
			}
			\If{$(\mi{x}>0) \wedge (i-d \geq 1) \wedge (L[t_{i-d}] < p')$ \label{ln:dBefore}}{
				$\mi{loadShifted} \gets \min(\mi{x},\mi{p'}-L[t_{i-d}])$\;
				$\mr{shift}(\mi{loadShifted}, L[t_i], L[t_{i-d}])$\;
				$\mi{x} \gets \mi{x} - \mi{loadShifted}$
			}
			$d \gets d+1$ \label{ln:shiftEnd}
		}		
	}
}
\Return $L$
\end{algorithm}
\vspace*{-1ex}

Algorithm~\ref{alg:parcut} receives input $c$ as the cut percentage 
and $L$ as the original load whose $\mi{PAR}$ is to be cut. 
We use Eq.~\ref{eq:cutPeak} to define the new target peak 
$p'$ (line~\ref{ln:newPeak}).
Then, for each time slot $t_i$, we verify 
whether the load at $t_i$ exceeds $p'$ (line~\ref{ln:check}). 
If this is the case, then we try to shift it (line~\ref{ln:shiftStart}-\ref{ln:shiftEnd}).
We first try to shift the load at $t_i$ to its neighboring time slot 
by incrementing variable $d$ for checking time slot 
$t_{i+d}$ (line~\ref{ln:dAfter}) 
and $t_{i-d}$ (line~\ref{ln:dBefore}).
However, when we have investigated all time slots
and there is still an amount of load to be shifted, 
then we conclude that it is not possible to do the cut, 
and the algorithm returns with failure (line~\ref{ln:fail}).

\vspace*{1ex}
\begin{example}[PAR-cut]
Let us assume that we have a set of time slots
$\mc{T} = \{t_1, t_2, \ldots, t_{24}\}$,
and load $L_{t_{19}}=5 \mi{kWh}$, 
$L_{t_{18}} = L_{t_{20}} = 2 \mi{kWh}$
and $L_{t_i}=1 \mi{kWh}$, 
where $1 \leq i \leq 17$, and $21 \leq i \leq 24$.
In this case we have $\mi{PAR}=4$.
Furthermore, let us assume that we want to cut 
the $\mi{PAR}$ by 40\% cut percentage.
Hence, our new peak $p'=3$ (see line~\ref{ln:newPeak}). 
First, Algorithm~\ref{alg:parcut} detects 
that we have an excess load at time slot 
$t_{19}$, and the excess load $x = 2 \mi{kWh}$.
Then, it proceeds with distributing 
the load to the neighboring time slot, $t_{20}$ and $t_{18}$. 
This makes $L_{t_{18}}=L_{t_{19}}=L_{t_{20}}=3 \mi{kWh}$.
At the end of this step, the excess load $x=0$, 
and the algorithm returns the newly modified load vector.
\end{example}
\vspace*{1ex}

For the proof of the soundness and the completeness 
of the algorithm, 
we refer to the input load to the algorithm as $L$, 
and the resulting load as $L'$.

\vspace*{1ex}
\begin{proof}[Soundness]
We need to show that if 
the algorithm returns $L'$, then this is correct 
(having property shown in Eq.~\ref{eq:fixLoadSum} and \ref{eq:cutPeak}).
%
%
Satisfying condition in Eq.~\ref{eq:fixLoadSum} is straightforward 
since we never decrease or add new load. 
The only modification we apply is shifting the load from one time slot to another.
Next, we always make sure that 
the maximum load in a time slot never 
exceeds $p'$ (see line~\ref{ln:check}). 
This makes the condition in Eq.~\ref{eq:cutPeak} hold.
\end{proof}

\vspace*{1ex}
\begin{proof}[Completeness]
We need to show that if the algorithm returns fail, 
then it is not possible to cut the load (as specified by the cut percentage $c$).
%
The only condition where the algorithm returns fail 
is when it reaches line~\ref{ln:fail}.
This means that we still have an excess load, and we have no slot left with load less than $p'$. 
Hence, in order to satisfy the condition in Eq.~\ref{eq:fixLoadSum}, we have to put this excess load in some time slot. But this will make this time slot have load greater than $p'$
which is not permitted by Eq.~\ref{eq:cutPeak}.
\end{proof}
%

\section{Multiunit Auction for Load Distribution}
\label{sec:multi}
In this section we explain how we run 
multiunit auction to distribute the cut load (the resulting load from Algorithm~\ref{alg:parcut}). 
The auction is run either in the beginning of the day or a day before.
It is held in several rounds until all loads are distributed.
We use $\mc{N}$ as the set of consumers 
and refer to agent $i$ as the bidding agent 
representing consumer $i\in \mathcal{N}$.

\subsection{Initial condition}
Let $L'$ denote the \emph{cut load vector} 
on which we are going to run the auction, 
and let $L$ denote the original load vector.
For a time slot $t \in \mc{T}$, 
whenever $L'(t) \geq L(t)$,
the demanded loads are distributed to the consumers.

Let $L_i^x$ be the \emph{actual load vector} 
for consumer $i \in \mc{N}$ 
(the actual loads delivered to consumer $i$'s residence) 
obtained in round $x$ of the auction and let round 0 
be the initial step before the auction begins.
Then, we have consumer $i$'s actual load: $L_i^0(t) = L_i(t)$ for $L'(t) \geq L(t)$ and time slot $t \in \mc{T}$.

\subsection{Minimal Load Guarantee}

It could be the case that at time slot $t$ there is a 
consumer $i \in \mc{N}$ who does not get any load because 
all loads have been won by other 
consumers. 
It means that at time $t$ consumer $i$ does not have any electricity.
This could happen when $L'(t) < L(t)$ for some $t \in \mc{T}$. That is, when we face a \emph{short of supply} condition (because of the cut) where the load supply at time $t$ is less than the actual consumers' demand. 

In order to prevent this condition from happening, 
we set $m$ as the minimal amount of load guaranteed for each consumer to have when we face a short of supply, i.e. for some time slot $t \in \mc{T}$, when $L'(t) < L(t)$:
\begin{enumerate}[(i)]
\item supply guarantee: $L'(t) \geq m \cdot |\mc{N}| $,  
\item minimal load guarantee: 
$L_i^0(t)=m$ for all $i \in \mc{N}$.

\end{enumerate}

\subsection{Initial Pricing}
\label{sec:initPrice}
To protect the energy company getting paid 0 
when we run the auction (this can happen when $W = \mc{N}$),
we define the reserve price per Wh 
for each time slot $t \in \mc{T}$ as 
\[p(t) = \frac{\mi{cost}(L(t))}{L(t)}.\]
\noindent
We define $\mi{cost}(L(t)) = c_1 L(t)^2+ c_2 L(t) + c_3$ 
(increasing and convex), such that 
time slots having higher load is more expensive
\cite{Mohsenian2010}.

\subsection{The Auction: Matching Demand Against Supply}

We use the multiunit auction as described in Section~\ref{sec:auction} with some modifications: 

\begin{enumerate}

\item 
An agent submits a set of bid 
$\{b_{1}, \ldots, b_{|\mc{T}|}\}$
all at once,
where $b_i$ is a bid for the load on time slot $t_i \in \mc{T}$.


\item Since it could be the case that 
not all loads are distributed to consumers 
on one round of the auction, 
we run multiple rounds of the auction where 
the loads available for the next round 
are the loads left from the current round. 

\item Bids (how much load an agent want 
and at what price) are placed simultaneously 
but separately for each time slot. 

\item The winners are determined for each time slot.

\item The auction is terminated when there is no 
load left or there is no bid submitted.

\end{enumerate}

\vspace*{0.5ex}
\noindent
\textbf{Initial condition. } 
Let $L^x(t)$ be the loads left (after it has been distributed to the winners) 
after the $x^{\mi{th}}$ round of the auction. 
Thus, we denote the initial loads available before the auction begins as 
$L^0(t) = L'(t) - \sum_{i \in \mc{N}} L_i^0(t)$ for time slot $t \in \mc{T}$, 
where round 0 is the initial step before the auction begins and $L'$ is the cut load vector.

\vspace*{0.5ex}
\noindent
\textbf{Bid. } 
In each round $x$ of the auction, agents place a bid 
on $L^{x-1}(t)$ using their own valuation
to get some load they desired 
for each time slot $t \in \mc{T}$. 
However, if the load available in $L^{x-1}(t)$ 
is less than what an agent needs, then she 
will place a bid on $L^{x-1}(t')$ where
$t'$ is the closest time slot to $t$ such that
$L^{x-1}(t')$ is greater than what she needs.
This is often called as \emph{myopic best response} strategy, i.e. each agent is interested 
to maximize her own utility given the current condition. 
%
%
Note that it is possible that a sophisticated agent 
might be able to manipulate the system by reasoning 
and acting over several time steps and do a bit better.
However, this involves sophisticated reasoning and
could well be too complex for more simple bidders.
And we will show that the optimal myopic strategy 
has attractive properties.


%

\vspace*{0.5ex}
\noindent
\textbf{Winners determination. }
The winners are determined separately for each time slot, 
i.e. we have a set of winners $W^x(t)$ for each time slot. 
Next, the resources available are updated to $L^x(t)$ by 
considering the load that has been distributed to the winners.

\vspace*{0.5ex}
\noindent
\textbf{Demand satisfiability. } Let $x^*$ be the final round of the auction.
In addition, for a consumer $i \in \mc{N}$
let $L'_i(t) = \sum_{x=0}^{x^*} L_i^x(t)$ be the sum of 
load she obtained through the auction for each time slot $t \in \mc{T}$.
Then, we require for consumer :
\vspace*{-2ex}
\[ \bigg( \sum_{t \in \mc{T}} L'_i(t) \bigg) = 
\bigg( \sum_{t \in \mc{T}} L_i(t) \bigg).\]
That is, the total load obtained by a consumer should be the same as her total electricity demand.

\vspace*{0.5ex}
\noindent
\textbf{Truthful bidding is a myopic best response. } 
We prove that the auction supports truthful bidding,
i.e. there is no incentive for agents 
to lie about their valuation.

\begin{proof}
Assume that there is still load that needs to be 
acquired by agent $i$ on round $x$.
Then, there are two cases: 

\begin{enumerate}
\item \label{case:1} 
$L^{x-1}(t)$ is equal or larger than 
what agent $i$ needs. 
Then, agent $i$ will bid on it. Assume that $v_i(t)$ is the agent $i$ valuation for the load in time slot $t$. There are two cases: 
	\begin{enumerate}
	\item \label{case:1.1} Agent $i$ does not win. 
	If agent $i$ bid an amount $< v_i(t)$, 
	it could not make her win. 
	However, if she bid an amount $> v_i(t)$,
	she could win. But then she has to pay 
	$v'$ (the bid of the highest non-winning bidder), 
	and it is also possible that $v' > v_i(t)$. 
	In this case, agent $i$ loss (because she has to pay 
	higher than her valuation of the electricity at that time).
	
	\item \label{case:1.2} Agent $i$ win. 
	Assume that the highest non-winning bid is $v'$. 
	This is also the price that the winners have to pay. 
	If agent $i$ bid $b'$, and $ v' \leq b < v_i(t)$, 
	then she still has to pay $v'$ 
	(she does not gain any additional benefit).
	However, if agent $i$ bid $< v'$, then she lost. 
	Hence, she does not get the load she want. 
	And if agent $i$ bid $> v_i(t)$,
	this also does not give any additional 
	benefit because she still has to pay $v'$.
	\end{enumerate}
	
\item $L^{x-1}(t)$ is less than what agent $i$ need. 
Then, agent $i$ choose the closest time slot to $t$, that is $t'$, 
such that $L^{x-1}(t')$ is equal or larger that what agent $i$ need. 
Then, the same the argument as in the two cases above (\ref{case:1.1} and \ref{case:1.2}), are applied here by replacing $t$ with $t'$.
\end{enumerate}
\end{proof}

\subsection{Consumer's Load Shifting}
Note that a bid for an amount of load at time slot $t \in \mc{T}$ placed 
by consumer $i$ does not necessarily comes from 
her actual need, from $L_i(t)$. 
Consider a simple example when 
consumer $i$ actually still needs to obtain her load for time slot 
$t' \in \mc{T}$ in the current round $x$, but $L^x(t') = 0$.
Since there is no more load available in time slot $t'$, 
then she has to place a bid for a different time slot $t'' \in \mc{T}$.
When she wins the bid, load shifting for consumer $i$ occurs, i.e. 
she shifts an amount of electricity consumption from time $t'$ to $t''$.

\subsection{Consumer's Utility}
\label{sec:utility}
We define consumer's utility after the auction  
by two terms:

\begin{enumerate}[(i)]

\item \emph{cost}: 
we compute the cost paid by a consumer at the end of an auction.

\item \emph{shift percentage}: 
in order to measure a consumer's inconvenience, 
we calculate the percentage of load that she has to shift.
The shift percentage of consumer $i$ is:
\[ \frac{\sum_{t \in \mc{T}} |L'_i(t) - L_i(t)|} {2 \cdot \sum_{t \in \mc{T}} L_i(t) }  .\]

\end{enumerate}


\section{Experiments}
\label{sec:exp}

In the experiments, we show 
the effects of the auction to consumer's utilities 
for different type of consumers.
In addition, we investigate whether 
there is any additional benefit 
for the consumer and/or the energy provider 
from running this auction.

\subsection{Experiment Setup}
\label{sec:setup}
In our experiments, we use an hourly time slot, i.e. $\mc{T} = \{1, \ldots, 24\}$, and total consumers, $|\mc{N}|=10000$. 
Then, we provide appliance usage setting to simulate consumers' electricity demand.
For the cost, any increasing and convex function can be used. 
In particular, we set 
$\mi{cost}(L(t)) = \big((L(t) + q_1) / (q_2 \sqrt{|\mc{N}|})\big)^2$ 
in order to include the energy company's marginal benefit, $q_1$,
and to control the price's growth, $q_2$ (we set $q_1=100$ and $q_2=1000$).

For consumer valuations (see Section~\ref{sec:auction}), we use $\alpha \cdot p(t)$, where $p(t)$ is the reserve price at time slot $t \in \mc{T}$ (see Section~\ref{sec:initPrice}). 
In order to set $\alpha$ among consumers, we use two types of distributions

\begin{enumerate}

\item \emph{US} distribution: for each consumer, 
her $\alpha$ is drawn from a set $\{1.0, 1.3, 1.5,1.6,1.9\}$ 
with distribution $0.4$, $0.2$, $0.2$, $0.1$, and $0.1$ respectively. 
This distribution is inspired by the US households wealth distribution in 2004
\footnote{http://www.faculty.fairfield.edu/faculty/hodgson/courses/so11/stratification\\/income\&wealth.htm
(accessed: 15 September 2012)}.

\item \emph{Uniform} distribution: for each consumer, her $\alpha$ is drawn uniformly from set $\{1.0, 1.1, 1.2 \ldots, 1.9\}$.

\end{enumerate}

Each experiment is run 10 times and
all results presented are with 95\% confidence interval. 
However, the intervals almost cannot be seen because the deviations are very small.

\subsection{System Cost}
\begin{figure}[tb]
\vspace*{-3ex}
\centering
\includegraphics[width=0.5\textwidth]{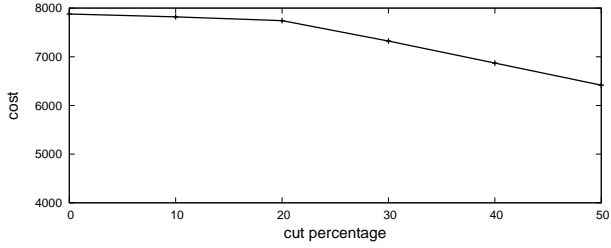}
\vspace*{-24ex}
\caption{
Overall system cost decrease as the cut percentage getting larger.
Simulation using 10000 households loads. 
}
\label{fig:system-cost}
\end{figure}

We present how the $\mi{PAR}$ cut affects 
the system cost.
We show the result up to 50\% cut percentage, 
since in our setting, it is not possible to cut the PAR with 60\% or more.
As expected, the more we cut the PAR, 
the lower the overall system cost is.
Our experimental results in Figure~\ref{fig:system-cost} 
shows that cutting the PAR up to 20\% does not give significant advantage.
However, cutting the PAR with 50\% cut percentage give almost 20\% overall cost reduction.
It looks very promising, but one have to be more careful and look on how consumer's utility changes with increasing cut percentage 
(as we will show later in this section).

\subsection{Consumers' Utility}
We measure consumers' utility from two perspectives. 
First, we look at how much they need to shift their
 consumption. Second, we look 
at the total cost that they need to pay.

\begin{figure}[tb]
\vspace*{-4ex}
\centering
\includegraphics[width=0.5\textwidth]{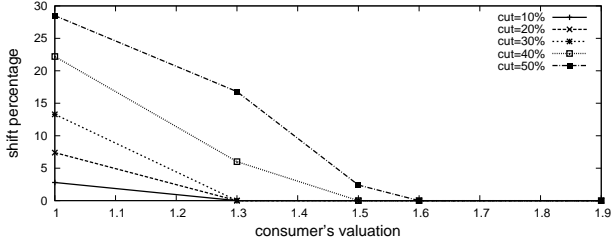}
\vspace*{-24ex}
\caption{
Consumers' utility based on shift percentage measurement (defined in Section~\ref{sec:utility}) grouped by their valuation (US distribution). 
}
\label{fig:shift-U}
\vspace*{-2ex}
\end{figure}

\begin{figure}[tb]
\vspace*{-3ex}
\centering
\includegraphics[width=0.5\textwidth]{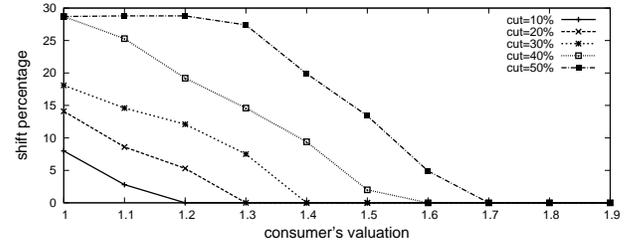}
\vspace*{-24ex}
\caption{
Consumers' utility based on shift percentage measurement (defined in Section~\ref{sec:utility}) grouped by their valuation (uniform distribution). 
}
\label{fig:shift-N}
\end{figure}

\vspace*{0.5ex}
\noindent
\textbf{Shift percentage. }
The results from both distributions (US and uniform) depicted in 
Figure~\ref{fig:shift-U} and \ref{fig:shift-N} shows the same trends, i.e.
consumers with lower valuation are the ones who shift the most.
This is expected since most of their demands which compete 
with higher valuation consumers are not satisfied (because they lost in the auction). Hence they need to shift their consumption (bid for another time slot).
For consumers with valuation more than $1.7$ we see no difference in their utility, 
and they do not shift at all. 
With this high valuation, we expect that they always win the auction and so see no difference in their supply.

\vspace*{0.5ex}
\noindent
\textbf{Cost paid. }
Figure~\ref{fig:shift-U} and \ref{fig:shift-N} both show 
the same trends that 
the higher the consumer valuation, the higher the cost.
This phenomenon is both caused by the reserve price and 
the auction itself.
Most of the consumers firstly will bid to satisfy their original consumption schedule, 
which happen most of the time in the peak time slot.
Although we have already cut the peak, 
the price of this time slot
is still the highest compared to the other 
(this is the time slot with the highest load). 
In addition, since most of the consumers bid on this time slot, 
most likely the winners for this time slot will be the 
higher valuation consumers, and the highest non-winning 
bid (which will determine the price they have to pay) 
will also be high.

As expected, low valuation consumers pay lower costs. 
This can also be seen as the tradeoff 
(from the previous results) 
that they need to shift more than the consumers with high valuation.
The setting with 50\% cut percentage offer lower cost paid compared to the others, but it also caused the low valuation consumers to shift more (see Figure~\ref{fig:shift-U} and \ref{fig:shift-N}).

\begin{figure}[tb]
\vspace*{-4ex}
\centering
\includegraphics[width=0.5\textwidth]{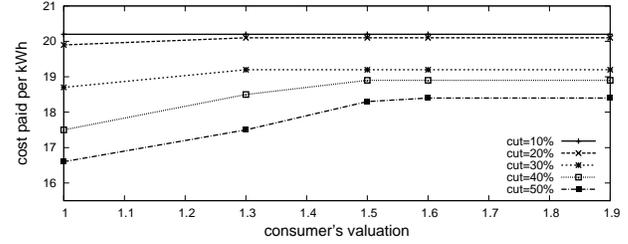}
\vspace*{-24ex}
\caption{
Consumers' utility based on the total cost paid 
(defined in Section~\ref{sec:utility}) grouped by their valuation (US distribution). 
}
\label{fig:cost-U}
\vspace*{-2ex}
\end{figure}

\begin{figure}[tb]
\vspace*{-6ex}
\centering
\includegraphics[width=0.5\textwidth]{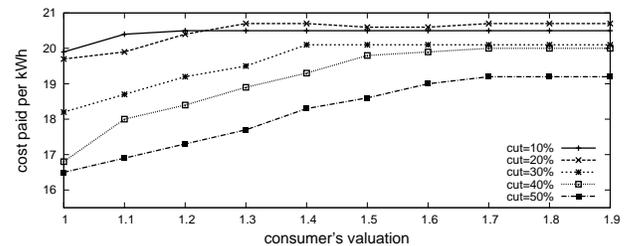}
\vspace*{-24ex}
\caption{
Consumers' utility based on the total cost paid 
(defined in Section~\ref{sec:utility}) grouped by their valuation (uniform distribution). 
}
\label{fig:cost-N}
\end{figure}

\subsection{Consumers' Additional Benefit}

\begin{figure}[tb]
\vspace*{-3ex}
\centering
\includegraphics[width=0.5\textwidth]{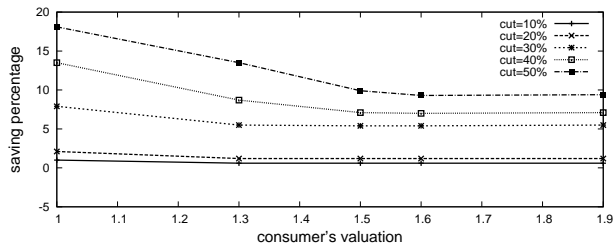}
\vspace*{-24ex}
\caption{
Consumers' cost saving percentage using the auction and PAR cut  
compare to the current system. Valuation used: US distribution.
}
\label{fig:saving-U}
\end{figure}

Compared to the current system, 
consumers gain an additional benefit using the PAR cut 
and the auction.
Consumers can save up to almost 20\% their electricity bill  
depending on the cut percentage implemented 
(see Figure~\ref{fig:saving-U}).
This happens without having to reduce the amount of 
their electricity consumption. 
For consumers with low valuation this saving 
can be seen as the tradeoff since they experience load shifting.
However, what is more interesting is that this saving also 
applies to consumers with high valuations 
(in this case, higher than $1.6$) 
who experience no load shifting at all (see Figure~\ref{fig:shift-U}).

\subsection{Company's Additional Benefit}
\begin{figure}[tb]
\vspace*{-4ex}
\centering
\includegraphics[width=0.5\textwidth]{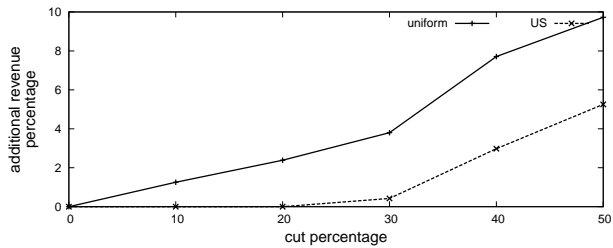}
\vspace*{-24ex}
\caption{
Company addition revenue from the auction.
}
\vspace*{-2ex}
\label{fig:company-add}
\end{figure}

In this experiment, we calculated the total cost paid 
by consumers and compared it to the energy company's 
cost model (the one that has to be paid by the consumers, 
as described in Section~\ref{sec:setup}).
Figure~\ref{fig:company-add} shows that  
the company experiences additional revenue by running 
the auction up to almost 10\% depending on the PAR cut implemented.
The larger PAR cut we have, the more peak loads are shifted. 
This will increase the price in the peak time slot because 
most likely the winners are consumers with high valuations, 
and the highest non-winning bid will most likely 
also be high.

\section{Conclusions and Future Work}
\label{sec:concl}

In this paper, we addressed 
the future vision of the smart grid to match
the demand with the supply.
Instead of relaying the PAR cut to the consumer, 
we cut PAR explicitly 
while still maintaining the same load.
Because the load available 
for each time slot are not necessarily 
the same as demanded by the consumers, 
we used a multiunit auction to distribute the load.
Our experiments showed that the overall system cost 
is decreasing as we have larger $\mi{PAR}$ cut percentages.
We also showed that there is a tradeoff 
between the price per kWh paid by a consumer
with the load shift percentage she experienced.
In addition, the consumer with low valuation 
normally has bigger shift percentage and lower price 
per kWh than the consumer with high valuation.
Another advantage is shown for the consumers that 
their total electricity cost is lower compare than their 
original electricity cost using the current setting (saving up to almost 20\%).
While for the low valuation consumers 
this can be understood as the tradeoff 
for their load shifting, 
the saving also happens for high valuation 
consumers who does not have any load shifting. 
%
The energy company also reaped 
an additional benefit from running the auction by 
acquiring up to almost 10\% additional revenue.
This phenomenon made the proposed approach look promising.

This work can be extended by considering 
a case when a consumer considers non-myopic strategies
and reasoning over multiple steps at once.
Our conjecture is that weak consumers 
(low valuation) can benefit from bidding on non-peak
time slots for her needs in peak time slots.
By bidding on non-peak time slots early,
a weak consumer has a better chance to avoid 
competition with stronger consumers 
(who also have to shift to those time slots 
because of even stronger consumers).
Furthermore, our approach can also be used
for load distribution when we have 
load fluctuation (for example, because 
of the renewable energy sources). 
Because the essence of our approach 
is to match the demand with the supply, 
we can use it either when we have to cut
the PAR, or when we have energy surplus 
to distribute.

\section*{Acknowledgment}
The research leading to these results has received funding from the European Union's Seventh Framework Programme (FP7/2007-2013) under grant agreement no: 288322.

\bibliographystyle{IEEEtranS}
\bibliography{matching-demand_wijaya}

\begin{thebibliography}{10}
\providecommand{\url}[1]{#1}
\csname url@samestyle\endcsname
\providecommand{\newblock}{\relax}
\providecommand{\bibinfo}[2]{#2}
\providecommand{\BIBentrySTDinterwordspacing}{\spaceskip=0pt\relax}
\providecommand{\BIBentryALTinterwordstretchfactor}{4}
\providecommand{\BIBentryALTinterwordspacing}{\spaceskip=\fontdimen2\font plus
\BIBentryALTinterwordstretchfactor\fontdimen3\font minus
  \fontdimen4\font\relax}
\providecommand{\BIBforeignlanguage}[2]{{%
\expandafter\ifx\csname l@#1\endcsname\relax
\typeout{** WARNING: IEEEtranS.bst: No hyphenation pattern has been}%
\typeout{** loaded for the language `#1'. Using the pattern for}%
\typeout{** the default language instead.}%
\else
\language=\csname l@#1\endcsname
\fi
#2}}
\providecommand{\BIBdecl}{\relax}
\BIBdecl

\bibitem{USDE06}
``Benefits of demand response in electricity markets and recommendations for
  achieving them,'' U.S. Dept. of Energy, Tech. Rep., 2006.

\bibitem{Bower2001}
J.~Bower and D.~Bunn, ``Experimental analysis of the efficiency of
  uniform-price versus discriminatory auctions in the england and wales
  electricity market,'' \emph{Journal of Economic Dynamics and Control},
  vol.~25, no. 3–4, pp. 561 -- 592, 2001.

\bibitem{Cappers10}
P.~Cappers, C.~Goldman, and D.~Kathan, ``Demand response in u.s. electricity
  markets: Empirical evidence,'' \emph{Energy}, vol.~35, no.~4, 2010.

\bibitem{Contreras2001}
J.~Contreras, O.~Candiles, J.~De~La~Fuente, and T.~Gomez, ``Auction design in
  day-ahead electricity markets,'' \emph{IEEE Transactions on Power Systems},
  vol.~16, no.~1, pp. 88 --96, Feb 2001.

\bibitem{Ganu2012}
T.~Ganu, D.~P. Seetharam, V.~Arya, R.~Kunnath, J.~Hazra, S.~A. Husain, L.~C.
  De~Silva, and S.~Kalyanaraman, ``nplug: a smart plug for alleviating peak
  loads,'' in \emph{Proceedings of the 3rd International Conference on Future
  Energy Systems: Where Energy, Computing and Communication Meet}, ser.
  e-Energy '12.\hskip 1em plus 0.5em minus 0.4em\relax New York, NY, USA: ACM,
  2012, pp. 30:1--30:10.

\bibitem{Leboudec11}
J.-Y. Le~Boudec and D.-C. Tomozei, ``Demand response using service curves,'' in
  \emph{IEEE PES - ISGT}, 2011.

\bibitem{Li2011}
N.~Li, L.~Chen, and S.~H. Low, ``Optimal demand response based on utility
  maximization in power networks,'' in \emph{Proceedings of the 2011 IEEE Power
  \& Energy Society General Meeting}, July 2011.

\bibitem{Mohsenian2010}
A.~Mohsenian-Rad, V.~Wong, J.~Jatskevich, R.~Schober, and A.~Leon-Garcia,
  ``Autonomous demand-side management based on game-theoretic energy
  consumption scheduling for the future smart grid,'' \emph{IEEE Transactions
  on Smart Grid}, vol.~1, no.~3, pp. 320 --331, Dec. 2010.

\bibitem{Oh2008}
H.~S. Oh and R.~Thomas, ``Demand-side bidding agents: Modeling and
  simulation,'' \emph{IEEE Transactions on Power Systems}, vol.~23, no.~3, pp.
  1050 --1056, Aug. 2008.

\bibitem{Ramchurn2012}
S.~D. Ramchurn, P.~Vytelingum, A.~Rogers, and N.~R. Jennings, ``Putting the
  'smarts' into the smart grid: a grand challenge for artificial
  intelligence,'' \emph{Commun. ACM}, vol.~55, no.~4, pp. 86--97, Apr. 2012.

\bibitem{Ramchurn2011}
S.~D. Ramchurn, P.~Vytelingum, A.~Rogers, and N.~Jennings, ``Agent-based
  control for decentralised demand side management in the smart grid,'' in
  \emph{The 10th International Conference on Autonomous Agents and Multiagent
  Systems - Volume 1}, ser. AAMAS '11.\hskip 1em plus 0.5em minus 0.4em\relax
  Richland, SC: International Foundation for Autonomous Agents and Multiagent
  Systems, 2011, pp. 5--12.

\bibitem{Strbac2008}
G.~Strbac, ``Demand side management: Benefits and challenges,'' \emph{Energy
  Policy}, vol.~36, no.~12, pp. 4419 -- 4426, 2008.

\bibitem{Vytelingum2011}
P.~Vytelingum, T.~Voice, S.~D. Ramchurn, A.~Rogers, and N.~R. Jennings,
  ``Theoretical and practical foundations of large-scale agent-based
  micro-storage in the smart grid,'' \emph{J. Artif. Intell. Res. (JAIR)},
  vol.~42, pp. 765--813, 2011.

\end{thebibliography}

\end{document}